\documentclass{article}
\usepackage{amsfonts,amssymb,amsmath}
\usepackage{graphicx}
\usepackage{graphics}
\newcommand{\R}{\mathbb{R}}

\begin{document}

\title{Ultrametric Model of Mind, I: Review}

\author{Fionn Murtagh \\
Department of Computer Science \\
Royal Holloway University of London \\
Egham, Surrey TW20 0EX, England \\
E-mail fmurtagh@acm.org}

\maketitle

\begin{abstract}
We mathematically model Ignacio Matte Blanco's principles of 
{\em symmetric} and 
{\em asymmetric being} through use of an ultrametric topology.
We use for this the highly regarded 1975 book of this 
Chilean psychiatrist and pyschoanalyst (born 1908, died 1995).
Such an ultrametric model corresponds to hierarchical clustering
in the empirical data, e.g.\ text.
We show how an ultrametric topology can be used as a 
mathematical model for the structure of the logic 
that reflects or expresses Matte Blanco's symmetric being,
and hence of the reasoning and thought processes involved in 
conscious reasoning or in reasoning that is lacking, perhaps 
entirely, in consciousness or awareness of itself.  
In a companion paper we study how symmetric (in the sense of 
Matte Blanco's) reasoning can be demarcated in a context of 
symmetric and asymmetric reasoning provided by narrative text.  
\end{abstract}

\section{Introduction}

Before introducing and then discussing in detail Matte Blanco's
work in section \ref{sect2},
in section \ref{ankh} we deal with alternative models of 
mental processes -- the geometry or the topology of mental 
processes.  These are neural model centered.   
Therefore they can be taken as influenced by physical models of 
the brain.  They provide a useful starting point for us because
they also avail of ultrametric topological representations and
processing frameworks.  

It will become clearer later that our primary
 motivation is not with a 
physical (and hence, let's assume, observable through 
physico-chemical and biological processes) view of the 
brain.  Instead our motivation is to have a processing
framework to model human reasoning that is, as reasoning, based on 
data such as text or dialog data, or other behavioral or expressive
signals. 

In section \ref{sect2} we survey Matte Blanco's ``logico-mathematical'',
albeit largely descriptive, 
theory of psychoanalysis. 

In section \ref{sect3} we look at how ultrametrics can provide
an appropriate framework for understanding Matte Blanco.  In
section \ref{sect4} we tie the ultrametric topology strongly 
to input data.  

In section \ref{sect5} we note how the unconscious can be vastly 
more efficient -- quicker in reasoning -- compared to the 
conscious mind.  An implication of this is why unconscious 
thought processes are of interest, as well as conscious 
reasoning, for computation and for decision making.

In general we address through our mathematical -- topological 
and, at times when metric, geometric -- modeling how the 
unconscious differs from the conscious.  Modeling for us 
consists of formulating and defining heuristic data structures,
with the intention of allowing us 
to explore how the unconscious can be expressed in terms of
measured data. 

\section{Topology of Mental Processes: Neuro-Cognitive 
Approach}
\label{ankh}

Anashin and Khrennikov (2009) present a modeling of cognitive 
processes and psychology, 
and point to how it is best to use the ultrametric topology 
viewpoint rather than the associated p-adic or m-adic (where p is 
prime, and m is positive integer) 
algebraic structure.   Mental entities are modeled as 
partially ordered balls
(hence sets) 
in physical space. 
``Mental topology is ultrametric'' (Anashin and Khrennikov, 2009, 
p.\ 486).  
This work pursues that started by Khrennikov 
at the end of the 1990s and described also in Khrennikov (2004).  

A distinction is drawn between the wiring of the brain,
a dendritic structure which is, 
as such, a hierarchical neuronal (possibly non-homogeneous) tree, 
on the one hand, and on
the other hand the ``mental trees'' of reasoning.  
Application is proposed to 
robotics, exemplifying behaviors that encompass certain types of
psychical behavior (``evolution of psyche of psycho-robots
(and even people interacting with them)'', p.\ 439).  


In one chapter (Anashin and Khrennikov, 2009, chapter 14, ``m-Adic 
modeling in cognitive science and psychology''), three models are 
described for cognitive systems with increasing sophistication 
as regards psychological and psychotic behaviors (such as neurosis, id\'ee fixe
and hysteria, but also emotions).  The dynamics of neural networks 
corresponding to the models are discussed.  In chapter 14
of Anashin 
and Khrennikov (2009), patterns of firing/non-firing neurons are 
studied, or frequencies of firing.  The longest common prefix or Baire metric, 
which is also an ultrametric, is used.  It is noted that these neural 
computation models, based on p-adic or m-adic 
dynamics (hence they comprise p-adic or m-adic neural networks) furnish 
geometrical models of psychology but the claim is not made of relevance
for neuro-physiology.  

The following chapter in Anashin and Khrennikov (2009) is entitled 
``Neuronal hierarchy behind the ultrametric mental space''.   By selecting
a priority node in a dendritic graph, rooted trees are determined based on 
either strings of firing neurons (giving rise to a ``mental point'') or 
based on patterns of spiking behavior.  Diffusion on these rooted trees 
leads to 
establishing probability models on collections of trees.  The mental 
activity in this case is based on a unit given by ``the state of a 
hierarchic neural pathway''.  Furthermore, ``Each psychological function 
is based on a hierarchic tree of neural pathways'', i.e.\ neuronal trees.
Mental encoding of information in this model
uses accounting of ``frequencies of firings
of neurons along the hierarchic neural pathways''.


\section{Conscious Reasoning and Sub-Conscious 
Thought Processes: Towards a Cognitive Model that Embraces Both}
\label{sect2}

In the previous section, models of mental processes were 
used.  These models were based on, for example, 
``mental points'' and neural pathways.  Applications, as 
we noted, can be to robotics including robotic behavior 
-- psychotic robotic behavior for example can be understood.  

Here in this article, our focus is more restrictive in 
one sense in that thought processes, and reasoning in 
particular, are at issue.  Rather than robotics as an 
application for us, instead we are interested in 
psychoanalysis and in particular text, literary and 
other bases for empirical data analysis and signal 
processing.  

In other work (e.g.\ Murtagh et al., 2009)
we have taken discussion of quality of artistic 
creation and studied how well this could be understood
computationally, i.e.\ through pattern recognition.
Analogously here, we find in the work of Matte Blanco 
an excellent basis for the statistical and computational 
study of human reasoning.  

In summary Matte Blanco associated the unconscious with 
``symmetries'' in thought processes; and he associated the 
conscious with ``asymmetries'' in reasoning.

\subsection{Matte Blanco's Psychoanalysis: A Selective Review}
\label{sect1}

Matte Blanco's {\em The Unconscious as Infinite Sets} (originally 
published in 1975; see Matte Blanco, 1998) was, according to the 
author, ``written for 
psycho-analysts as well as for mathematical philosophers'' 
and is
described in Eric Rayner's Foreword as ``undoubtedly [his] most fundamental
work''.  

We will begin by summarizing particularly salient aspects
of Matte Blanco's work under the points laid out as follows.  That 
will help in showing how the various points can be 
seen in mathematical terms.  Quotations in the following are 
from Matte Blanco (1998).  

\begin{enumerate}

\item Matte Blanco related his work to Freud's conscious or unconscious.

\begin{itemize}
\item Relative to Freud's work, Matte Blanco had it 
``largely reformulated in terms of symmetry and asymmetry''.  
\item For him, these were ``two kinds or modes of being rather 
than of existence''.   
The interplay of symmetry and asymmetry is the focus of 
Matte Blanco's work.
\item The upshot of this was that Matte Blanco arrived at what
he termed a bi-logical system or bi-logic.  
\end{itemize}

There are ``two fundamental types of being which exist within the unity of 
every man: that of the `structural' id (or unrepressed unconscious or system
unconscious or symmetrical being) which becomes understandable with the help
of the principle of symmetry; and that visible in conscious thinking, which 
can roughly be comprehended in Aristotelian logic.''  

Freudian consciousness and unconsciousness are reformulated in terms of 
symmetrical and asymmetrical modes of being.  
It is to be noted that this is 
not a Freudian ``rational-irrational'' polarity 
but rather, on the side of the symmetric mode of being, the 
``unrepressed unconscious'', or what is ``the unconscious by its 
own nature or structural unconscious''.  
As seen in the development of the theory of Matte Blanco, 
``It is an attempt at putting in 
logico-mathematical terms the findings of Freud''. 

\item
Symmetrization is a principle which, as shown in Matte Blanco, 
helps in understanding:

\begin{itemize}
\item schizophrenia, and clinical treatment and practice
\item metaphor and other figures of speech
\item jokes, and disjunctions or abrupt change in discourse
\item emotion and emotionally loaded language
\item dream
\item poetry, literature, art
\item subconscious
\item the ``structured'' id (or unrepressed unconscious)
\item our system unconscious (or symmetrical {\em being})   
\end{itemize}

What is especially considered by him is ``pure symmetry'' 
at the deepest level, i.e.\ at ``the level of {\em being}, in 
contrast to the level of {\em happening}'' (emphasis in original).

For Matte Blanco, ``Symmetrical being is the normal state of man.''  
From it, its counterpart (in a way, its dual), 
``consciousness or asymmetrical being emerges'' 
and ``makes attempts at describing it'' (i.e.\ the primary 
experience of symmetric being).   
Justifiably we can consider ``symmetrical logic'' 
as the framework
of this description by asymmetrical thought of primary symmetrical 
thought.  
He says: 
``the most central trait'' of symmetrical being ``is the peculiar
(extensive) use of symmetrical relations'' -- hence, ``the 
symmetrical mode of being or symmetrical mode.''  

\item 
Within a class of things as conceptualized by the thinking 
person, there is perfect equivalence of class members, implying the
following. 

\begin{itemize}  
\item no contradiction 
\item absence of negation 
\item displacement 
\item space and time vanish 
\item no relations of contiguity 
\item arising from the last-mentioned: no order 
\end{itemize}

\item How a class is defined in practice, or is known to the 
thinking person, is described in these terms.  

\begin{itemize}
\item Because, as elaborated on in Matte Blanco, one class member 
is -- in terms of class membership -- indistinguishable from 
another class member, we have the following:
``the unconscious does not know individuals but only classes or 
propositional functions which define the class''.  

Further, 
``The only unity for the (symmetrical) unconscious is the class 
or set, in which all individuals belonging to it are included.
The unconsciousness cannot, therefore, deal with parts, except 
by treating them as classes or sets.''  

\item 
``Consciousness ... when confronted by a whole class can only 
consider it in two ways: either it focuses on the limits (or 
definition) of the class, that is, on those precise features 
which characterize it and distinguish it from all other classes,
or it concentrates on the individuals which form the class.''  
\end{itemize}

\item A class comes about through condensation.
\begin{itemize}
\item ``... two impulses which appear incompatible in Aristotelian
logic and their union in one expression, ... is accomplished in 
condensation''   
\end{itemize}

\item The principle of generalization relates different classes.  
\begin{itemize}
\item We assume various classes. 
\item Then ``the principle of generalization and the 
principle of symmetry'' 
are both taken for their explanatory capability in regard to classes.
\item In this way, the ``generalizing part [in the human] leads to 
symbols'', since symbols arise out of knowledge of, or awareness of, 
classes.  
\item Classes are structured as, what might be called, ``bags of symmetry'' 
(in quotation marks in the original), 
and also ``levels''.  

\end{itemize}

\item Counterposed to the symmetrical principle in Matte Blanco is the 
asymmetrical principle. 
\begin{itemize}
\item It is visible in conscious thinking.  
\item It can roughly be comprehended in, or expressed through, 
Aristotelian logic.
\item 
``Asymmetrical being ... perceives reality as divisible 
or formed by parts and, 
as such, related to spatio-temporality''. 
\item Symmetrical being can by known only through the glass or prism 
of asymmetrical being: ``Thinking requires asymmetrical relations.  So does
consciousness.''   
\end{itemize}

\item Quantifying the symmetrical.

\begin{itemize}
\item ``Symmetrical being alone is not observable in man.''  
Even delineating it is ``already an asymmetrical ... activity''.  

\item In regard to emotion, the 
``magnitude of emotion'' is understood in terms of ``the proportion between
symmetrical and asymmetrical thinking''.   
\item 
``[U]nconscious psychological events are not intrinsically 
immeasurable'' 
although compared to a physical event being susceptible to 
just one measurement, instead with unconscious events it is 
a matter of being susceptible to infinite measurement -- 
understood on the basis of the Cantor argument whereby 
a whole set, being in a bijection with a part of this same set, 
implies the same countable infinite cardinal for both whole and part 
sets.   

\item ``By making the individual identical to the class, 
the principle of symmetry, {\em as seen from an asymmetrical point
of view}, leads to the infinite set ...''  

\item ``We must ... keep in mind the possibility that if 
things are viewed in terms of multidimensional space, symmetrical 
being can actually unfold into an infinite number of asymmetrical
relations.''   

\end{itemize}

\item In free recall, and in other areas besides such as 
in literature, words are tracers for expressing what 
lies behind.  

\begin{itemize}
\item 
``Consciousness cannot exist without asymmetrical relations, because the 
essence of consciousness is to distinguish and to differentiate and 
that cannot be done with symmetrical relations alone.''  

\item ``Symmetrical being is translated into asymmetrical terms 
by means of words.  {\em Words (i.e.\ their meanings) are the 
asymmetrical tools of the translating-unfolding function.}''
(Italics in the original.)

\item We have that ``words, abstract things, fulfill the function of 
differentiating between concepts and also between other things.  They 
are bound to be, therefore, highly asymmetrical in their structure.'' 

\item To the foregoing we can add:
Text is the ``sensory surface'' (McKee, 1999, formulated in statistical and
computational terms in Murtagh
et al., 2009) of the underlying semantics.
In section \ref{secttext} we will return to 
further motivation as to why words are a good starting point 
for further analysis and how this can even go towards accessing
aspects of underlying symmetrical being.  
\end{itemize}

\end{enumerate}

Thus far, we have selected various central themes from Matte Blanco.  
This leads us to a conclusion drawn by Lauro-Grotto (2007) that directly 
follows from Matte Blanco: 
``... here comes my observation: the structural unconscious, in the 
way it is reformulated by Matte Blanco, the symmetric mode -- all this 
is homologous to an ultrametric structure. The
generalization principle reflects the hierarchical arrangement in which 
all the stimuli (or concepts) are perceived as belonging to classes, and 
the classes are clustered into super-classes of increasing generality. 
Finally, a single omni-comprehensive class is generated.''  

In a word, an ultrametric topology means that all that we are dealing
with is to be found in a hierarchy or a tree structure.   

Lauro-Grotto (2007) points to how equi-similar (or equi-distant)
stimuli or concepts indicate an ultrametric (or hierarchy, or tree) topology.  
In this work, we will go further.  

We will show how the laying out 
by Matte Blanco of the symmetric and asymmetric principles leads 
in a very natural way to an ultrametric topology as a representational
model.  An ultrametric 
space is defined by two of the four possible triangle configurations 
in a Euclidean space, viz.\ 
that they be either isosceles with small base, or equilateral. 

The isosceles with small base case does not detract one iota from 
symmetry.  Murtagh (2009) explores the many ways and contexts in which 
a hierarchy expresses symmetry.  There is a huge advantage for us in 
considering especially the isosceles with small base case of ultrametricity: 
it models novelty, or anomaly, or change.  We will illustrate that below
(in section \ref{sect4}).

\section{Ultrametric Topology, Background and Relevance}
\label{sect3}

Having surveyed Matte Blanco's view of unconscious thought processes
expressed as (Matte Blanco's term) symmetry, and conscious reasoning
expressed as (again Matte Blanco's term) asymmetry, in this section
we will lay out a basis for mathematically modeling these -- 
symmetry, asymmetry -- as respectively ultrametric (i.e.\ metric on 
a tree or hierarchy) and metric.  

As observed by Lauro-Grotto (2007, p.\ 539), 
the aspect of anomaly modeling 
via an ultrametric is nicely consistent with Matte Blanco's symmetrical
logic:
``... we know that something similar can actually be experienced in 
finite space when we look at a very distant three-dimensional structure 
and we perceive it as though it were a single point. Symmetrization of 
relationships can therefore be described as a transition from a metric 
to an ultrametric conceptual organization.'' 

\subsection{Metric}

\begin{figure}
\begin{center}
\includegraphics[width=6cm]{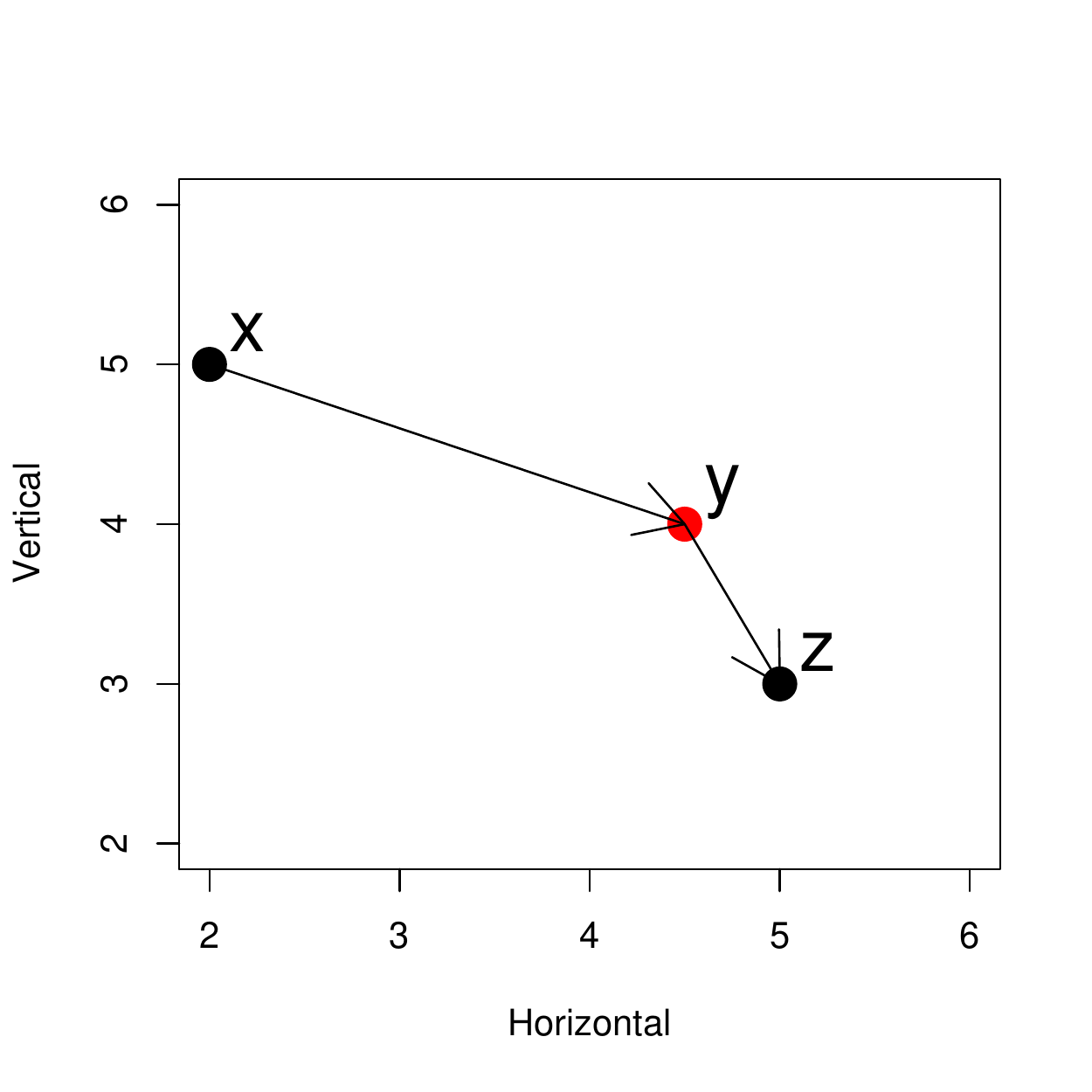}
\end{center}
\caption{The triangular inequality defines a metric:
every triplet of points satisfies the relationship:
$d(x,z) \leq d(x,y) + d(y,z)$ for distance $d$.}
\label{fig2}
\end{figure}

The triangular inequality holds for metrics.  An example of a metric is
the Euclidean distance, illustrated in Figure \ref{fig2}, where each
and every triplet of points satisfies the relationship:
$d(x,z) \leq d(x,y) + d(y,z)$ for distance $d$.  Two other relationships
also must hold.  These are symmetry and positive definiteness, respectively:
$d(x,y) = d(y,x)$, and $d(x,y) > 0 $ if $x \neq y$, $d(x,y) = 0 $ if $x = y$.

Semantic analysis based on a metric embedding is pursued 
in Murtagh (2005a), especially chapter 5 dealing with many types of 
textual content, including technical, literary, and philosophy.   
In Murtagh et al.\ (2009), in an extensive 
analysis of 
film script, it is shown how emotion can be traced out, and this is 
achieved in an unsupervised way from the text input alone.  

\subsection{Ultrametric}

\begin{figure}[t]
\begin{center}
\includegraphics[width=6cm]{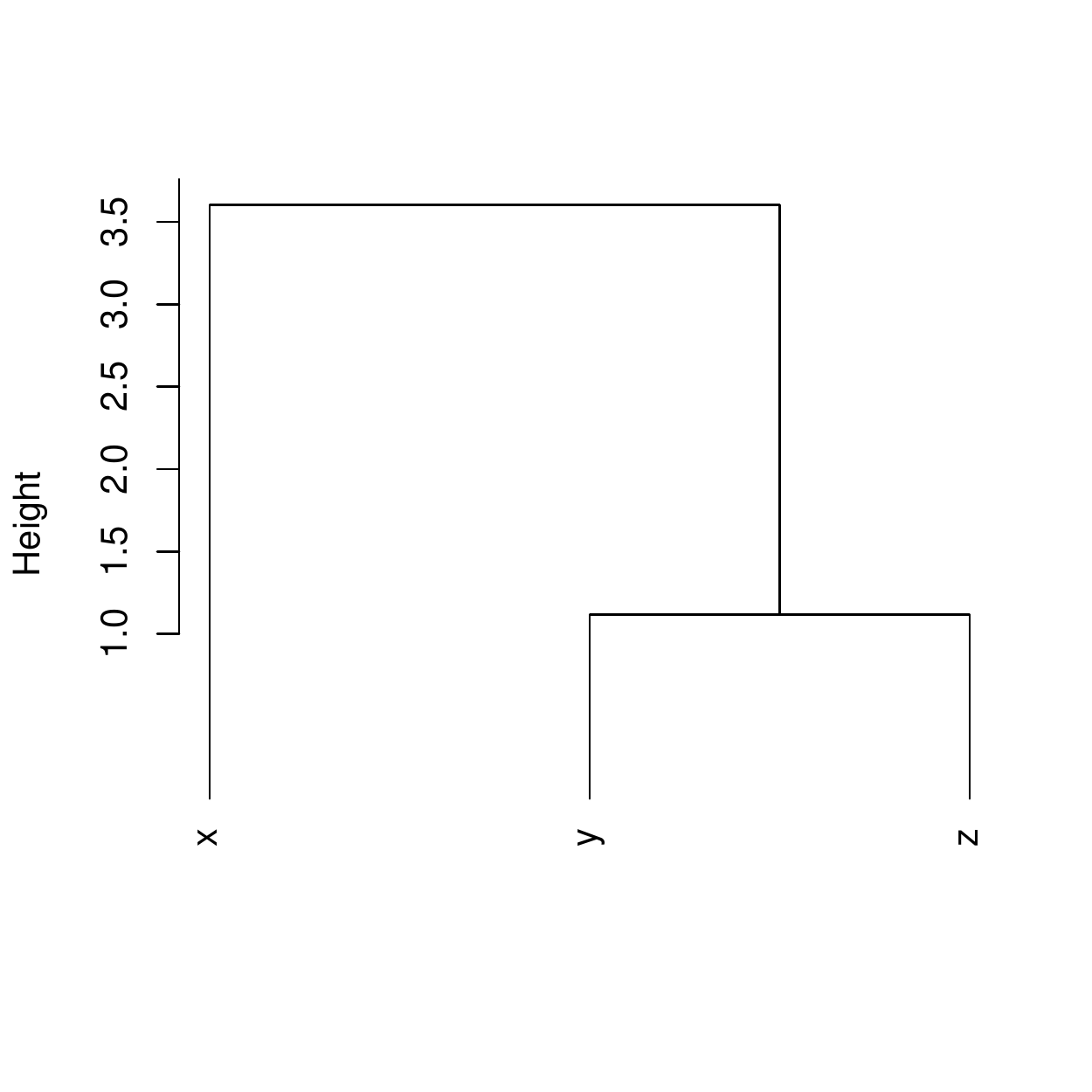}
\includegraphics[width=6cm]{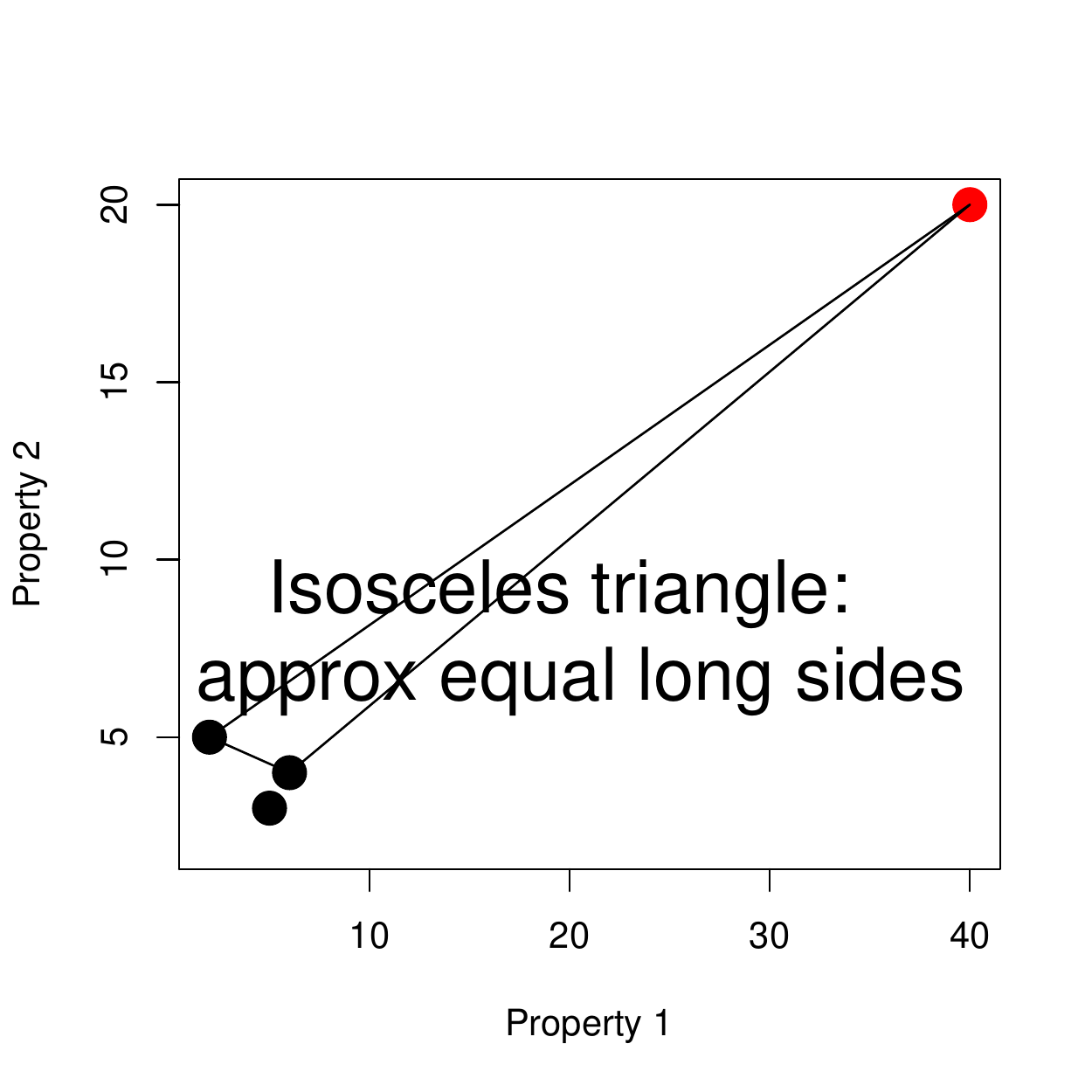}
\end{center}
\caption{Left: The strong triangular inequality defines an ultrametric:
every triplet of points satisfies the relationship:
$d(x,z) \leq \mbox{max} \{ d(x,y), d(y,z) \}$ for distance $d$.
Cf.\ by reading off the hierarchy, how this is verified for all $x, y, 
z$. In addition the symmetry and positive definiteness conditions
hold for any pair of points.
Right: The ``new arrival'' is on the far right.  While we can easily
determine the closest target (among the three objects represented
by the dots on the lower left), the {\em approximate} closest target is 
ambiguous.  This motivates the ultrametric as opposed to the metric 
viewpoint.}
\label{fig4}
\end{figure}

An ultrametric, compared to a metric, requires a stronger 
relationship between all triplets of points.  The ultrametric is illustrated 
in Figure \ref{fig4}, left.

To see how an ultrametric is a good mathematical model of anomaly, or
exception, or novelty, consider Figure \ref{fig4}, right.  Say, for 
example, there is the situation of seeking best match material, and 
our search term and the existing material are shown as points in Figure
\ref{fig4}, right.   If the target
population has at least one good match that is close to the query,
then this is (let us assume) clearcut.  However if all matches in the
target population are very unlike the query, does it make any sense to
choose the closest?  Whatever the answer here we are focusing on the
inherent ambiguity, which we will note or record in an appropriate way.
Figure \ref{fig4}, right, illustrates this situation where
the query is the point to the upper right.  Relative to the illustration
in Figure \ref{fig4}, left, the query point would be associated with 
terminal node, x. 

Note that our illustration in Figure \ref{fig4}, right, uses 
approximate similarity of the long sides of the triangle shown.  
This is by way of background for the companion paper, Murtagh (2012).  

\section{From Data to an Ultrametric: Hierarchical Clustering} 
\label{sect4}

%
%
%
%

\subsection{Ultrametric Topology and Hierarchy}

The ultrametric topology was introduced by Marc Krasner (1944),
the ultrametric inequality having been formulated by Hausdorff in 1934.
 
Essential motivation for the study of this area is
provided by Schikhof (1984, see in particular chapters 18, 19, 20, 21) 
as follows.  Real and complex fields gave rise
to the idea of studying any field $K$ with a complete valuation $| . |$
comparable to the absolute value function.  Such fields satisfy the
``strong triangle inequality'' $| x + y | \leq \mbox{max} ( | x |,  
| y | )$.  Given a valued field, defining a totally ordered Abelian
(i.e.\ commutative) group,
an ultrametric space is induced through $| x - y | = d(x, y)$.

Various terms can be used interchangeably for analysis in and
over such fields such as p-adic, ultrametric, and non-Archimedean.

The natural geometric ordering of metric valuations is on the real line,
whereas in the ultrametric case the natural ordering is a hierarchical
tree.

\subsection{Further Properties of Ultrametric Spaces}
\label{propum}

As already noted, in 
an ultrametric space all triangles are either isosceles with
small base, or equilateral.  We have here very clear symmetries of
shape in an ultrametric topology.   These symmetry ``patterns'' can be
used to fingerprint data sets and time series: see
Murtagh (2004, 2005b) and Ezhov et al.\ (2008) for many examples of this.

Some further properties that are studied in Lerman (1981, chapter 0, 
part IV), Chakraborty (2005)
and van Rooij (1978) are as follows. 

\begin{enumerate}

\item Every point of a circle in an ultrametric space is a center of the circle.
\item In an ultrametric topology, every ball, i.e.\ cluster, 
is both open and closed (termed {\em clopen}).
\item An ultrametric space is 0-dimensional.
\end{enumerate}

It is clear that an ultrametric topology is very different from our
intuitive, or Euclidean, notions.  The most important point to keep
in mind is that in an ultrametric space everything ``lives'' in a
hierarchy expressed by a tree.

The fact that every member of a class can be taken as the center of a 
circle points to a strong form of equivalence of the class members.   
Lerman (1981) and Chakraborty (2005) discuss this property.

The property of being both topologically open and closed is due to 
these properties of a set not being contraries.  Rather, 
a closed set is such that its complement is open.  The set can also be 
open by virtue of being a union of open sets.  Open and closed are 
simultaneous properties in this case.  It is important to note that
were we to take our cluster members (concepts or whatever) on the
real number line, then the context would be different.  

The clopen property lends itself well to a class being defined either 
by what it is not, or being defined through it being a union of 
other classes.  

A 0-dimensional space is studied in van Rooij (1978) or  
Chakraborty (2005).  If the space is constituted of clopen sets --
there is a countable base or covering of clopen sets -- then 
it is 0-dimensional.  

The dimensionality of a 
collection of points is 0.  So, in the present 
context and informally expressed, 
to say we have a 0-dimensional space is the same as saying 
that each cluster or class we are dealing with will always be akin 
to a point.  That is also to claim that the members of the class, 
for our purposes here, are tantamount to being identical.  

To conclude, therefore, we have sought to show how 
well an ultrametric space models 
Matte Blanco's symmetry, as surveyed in section \ref{sect1}, 
and how ultrametric space provides a framework for 
understanding symmetrical being, or a mathematical model of symmetrical 
being.   

%

\subsection{Inducing an Ultrametric through Agglomerative 
Hierarchical Clustering}

Inducing an ultrametric means, in practice, building a hierarchical 
clustering from given data.  

A mapping of metric to ultrametric is achieved by an agglomerative
hierarchical clustering algorithm, a well-established approach that 
depends on a cluster (compactness, or connectedness, or other) criterion.
If each of $n$ observations is taken as a singleton node in the hierarchical 
tree, $n-1$ pairwise agglomerations take place, producing a new node in 
the tree with each agglomeration.  There is redefinition of inter-cluster 
distance (or, less restrictive, dissimilarity) following each agglomeration.
Such a tree is 2-way or binary.   A p-adic encoding of observations 
can be read off (Murtagh, 2009, 2010b).  

The agglomerative algorithm sketched out here is improved in 
practice, for computational efficiency reasons. Instead nearest neighbor
chains of cluster centers are created, and updated following agglomerations
which take place whenever a reciprocal nearest neighbor pair is found.  
See Murtagh (2005a) and references therein.  

One agglomerative clustering criterion is a minimal connectedness one,
giving rise to the single link hierarchical clustering method.  It can be
shown that the resulting ultrametric between any pair of points (or 
observations) is less than or equal to the starting distance between this 
pair of points.  For this reason it is termed the sub-dominant ultrametric.  
Rammal et al.\ (1986) use the sub-dominant ultrametric.  

Finally, in this short discussion of hierarchical clustering, to 
draw yet another link to the work of Matte Blaco, it is noted in 
Rayner's (1995) review of Matte Blanco that the latter's 
investigation of ``{\em process of thinking} ... emphasizes the essential
centrality of {\em classificatory} activity at all levels of thought,
even in the unconscious.''  
This is a useful background consideration for the introduction 
to empirical and quantitative data analysis in  section \ref{secttext}
below.  



\subsection{Short Review of Hierarchical Clustering Algorithms}

Agglomerative hierarchical clustering has been the dominant
approach to constructing embedded classification schemes.
It is often helpful to distinguish between {\em method}, involving a
compactness criterion and the target structure of a 2-way tree
representing the partial order on subsets of the power set; as
opposed to an {\em implementation}, which relates to the detail
of the algorithm used.

As with many
other multivariate techniques (i.e., input data consists of 
measures on an object set crossed by an attribute set, so the 
obects are said to be multivariate), 
the objects to be classified have
numerical measurements on a set of variables or attributes.
Hence, the  analysis is carried out on the rows of an array or matrix.
If we do not have a matrix of numerical values to begin with, then
it may be necessary to  construct such a matrix from qualitative or 
symbolic data.
The objects, or rows of the matrix, can be viewed as vectors in
a multidimensional space (the dimensionality of this space being
the number of variables or columns).  A geometric framework of
this type is not the only one which can be used to formulate
clustering algorithms.
Suitable alternative forms of storage of a rectangular array of
values are not inconsistent with viewing the problem in geometric
terms (and in matrix terms -- for example, expressing the
adjacency relations in a graph).

Motivation for clustering in general, covering hierarchical clustering
and applications, includes the following:
analysis of data; interactive user interfaces; storage and retrieval;
and pattern recognition.

Surveys of clustering with coverage also of hierarchical clustering
include  Gordon (1981), March (1983), Jain and Dubes (1988), Gordon (1987),
Mirkin (1996),
Jain et al.\ (1999), and Xu and Wunsch (2005).
Lerman (1981) and Janowitz (2010) present overarching
reviews of clustering including through use of lattices that generalize
trees.
The case for the central role of hierarchical clustering in information
retrieval
was made by van Rijsbergen (1979) and continued in the work of Willett
(e.g.\ Griffiths et al., 1984)
and others.  Various mathematical views of hierarchy, all expressing
symmetry in one way or another, are explored in Murtagh (2009).

\subsection{Ultrametrics and Logic}

The usual
ultrametric is an ultrametric distance, i.e.\ for a set I,
$d: I \times I \longrightarrow \R$
(so the ultrametric distance is a real value).  The generalized
ultrametric is:
$d: I \times I \longrightarrow \Gamma$,
where $\Gamma$ is a partially ordered set.  In other words, the
{\em generalized} ultrametric distance is a set.  With this set one
can have a value,
so the usual and the generalized ultrametrics can amount to more or
less the same in practice (by ignoring the set and concentrating on its
associated value).  After all, in a dendrogram one does have
a set associated with each ultrametric distance value (and this is most
conveniently the terminals dominated by a given node; but we could have other
designs, like some representative subset or other,
of these terminals).  Most usefully the set,
$\Gamma$, is defined from the original attributes, which we denote by the
set $J$; whereas the sets of observations read off a dendrogram are
subsets of the observation set (which we label with the index set $I$).
So $\Gamma = 2^J$ (and not $2^I$).

In the theory of reasoning, a  monotonic operator is rigorous application
of a succession of  conditionals (sometimes called consequence
relations).  However:
``In order to deal with programs of a more general kind (the 
so-called disjunctive programs) it became necessary to consider
multi-valued mappings'', supporting non-monotonic reasoning in the
 way now to be described (Priess-Crampe and Ribenboim, 1999, pp.\ 10, 13).
The novelty in the work of Priess-Crampe and Ribenboim (1999, 2000)
is that these
authors use the generalized ultrametric as a multivalued mapping.
(A more critical view of the usefulness of the generalized ultrametric
perspective is presented by Kroetzsch, 2006). 

The generalized ultrametric approach has been motived (Seda and 
Hitzler, 1998)
as follows.  ``Situations arise ... in computational logic in the presence 
of negations which force non-monotonicity of the operators involved''.
To address non-monotonicity of operators, one approach
has been to employ metrics in studying
some problematic logic programs.  These ideas were taken further in
examining quasi-metrics, and
generalized ultrametrics i.e.\ ultrametrics which take values
in an arbitrary partially ordered set (not just in the non-negative
reals).  Seda and Hitzler (1998) 
``consider a natural way of endowing Scott
domains [see Davey and Priestley (2002)]   
with generalized ultrametrics.  This step provides a technical
tool [for finding fixpoints -- hence for analysis] of non-monotonic
operators arising out of logic programs and deductive databases and
hence to finding models for these.''

A further, similar, viewpoint is (Seda and Hitzler, 2010): 
``Once one introduces negation, which is certainly implied by
the term {\em enhanced syntax} ... then certain of the important
operators are not monotonic (and therefore not continuous), and in
consequence the Knaster-Tarski theorem [i.e.\ for fixed points; 
see Davey and Priestley, 2002] 
is no longer applicable to them.  Various ways have been proposed to
overcome this problem.  One such [approach is to use] syntactic conditions on
programs ... Another is to consider different operators ... The
third main solution is to introduce techniques from topology and
analysis to augment arguments based on order ... [latter include:]
methods based on metrics ... on quasi-metrics ... and finally ...
on ultrametric spaces.''

The convergence to fixed points that are based on a generalized
ultrametric system is precisely the study of
spherically complete systems and expansive automorphisms.
See Murtagh (2009) for a short introduction.

\section{Norm-Referenced Reasoning and Unconscious Thought Processes, 
Contrasted with Prototype-Referenced Reasoning and Conscious Reasoning}
\label{sect5}

In Murtagh (2010b) we develop a generative theory of information.  
Given that algorithmic complexity views the complexity of an 
object as the work required to generate it, we characterize the
work needed to generate an object in an ultrametric space 
as ultrametric algorithmic information.   
The approach uses a hierarchy as a ``key'' to the generative mechanism 
for an object.  It is a {\em norm-referenced} approach. 

This leads to further support for the hierarchical model, hence an 
ultrametric topology, as a model for unconscious thought, as we will
now discuss.  

In Giese and Leopold (2005), it is found that norm-referenced 
encoding of human
faces is a more likely mechanism in facial recognition, compared to
example-based encoding.  The former is with
reference to an average or norm, whereas the latter is relative to
prototypical faces.  

Leopold et al.\ (2006) reinforce 
this: ``The main finding 
was a striking tendency for neurons to show tuning that appeared 
centered about the average face''.  They suggest that norm-referencing
is helpful for making face recognition robust relative to viewing
angle, facial expression, age, and other variable characteristics.
Finally they suggest: ``Norm-based mechanisms, having adapted to our 
precise needs in face recognition, may also help explain why our 
[human] face 
recognition is so immediate and effortless...''

A wide range of experimental psychology results are presented by
Dijksterhuis and Nordgren (2006) 
to support the link between norm-referenced reasoning and
unconscious reasoning, on the one hand,
contrasted with the link between prototype-referenced
reasoning and conscious thinking, on the other hand.  We will pursue some
discussion of these links since they provide a most consistent backdrop to our
work.

Encoding of information is fundamental.
``Thinking about an object implies that the representation of 
that object in memory changes.''  Furthermore,
``information acquisition'' remains crucial for either
form of thought, conscious or unconscious.

Dijksterhuis and Nordgren (2006) point to how conscious thought
can process between 10 and 60 bits per second.  In reading, one processes about
45 bits per second, which corresponds to the time it takes to read a fairly 
short
sentence.  However the visual system alone processes about 10 million bits per
second.  It is concluded from this that the conscious thinking process in
humans is very low, compared to the processing capacity of the entire human
perception system.
 
We advance here a hypothesis as to why human thinking includes
unconscious as well as conscious thought.  Namely, we note that
{\em conscious} reasoning is slow compared to the vastly more 
efficient and dramatically faster processing speed of {\em 
unconscious} thought processes.  
Dijksterhuis and Nordgren (2006) also point to how unconscious 
thought is less precise and carries no order, including chronological, 
information.  We have already noted these aspects in Matte Blanco's 
symmetry and our ultrametric interpretation (subsections \ref{sect1} 
and \ref{propum}). 

Conscious thought therefore is
both limited and limiting.  A small number of foci of interest
(``only one or two attributes'') have to
take priority.  There are inherent limits to conscious thought as a result.
 As a result of limited capacity,
``conscious thought is guided by expectancies and schemas''.
Limited capacity therefore goes hand in hand with use of stereotypes or schemas.
``... people use ... stereotypes (or schemas in general) under
circumstances of constrained processing capacity ...  [While]
this [gives rise to the conclusion] that limited processing capacity
during {\em encoding} of information leads to more schema use,
[current work proposes] that this is also true for thought
processes that occur after encoding.  ... people stereotype more
during impression formation when they think consciously compared
to when they think unconsciously.  After all, it is consciousness
that suffers from limited capacity.''

It may, Dijksterhuis and Nordgren (2006) proceed,
 be considered counter-intuitive that stereotypes are
applied in the limited capacity, conscious thought, regime.
However stereotypes may be ``activated automatically (i.e., 
unconsciously)'', but ``they are applied {\em while we 
consciously think} about a person or group''.
Conscious thought is therefore more likely to (unknowingly)
attempt ``to confirm an expectancy already made''.

On the other hand,
unconscious thought is less biased in this way, and more slowly
integrates information.  ``Unconscious thought leads
to a {\em better organization} in memory'', arrived at through ``incubation''
of ideas and concepts.
``The unconscious works ... aschematically, whereas consciousness 
works ...  schematically''.   ``... conscious thought is more like 
an architect, whereas unconscious thought behaves more like an archaeologist''.

Viewed from the perspective of the work discussed in this subsection, it can
be appreciated that our hierarchical and generative description of an object
set is a simple model of unconscious thought.  That it is simple is clear:
to begin with, it is static.   Our hierarchical and generative description of
an object set (cf.\ Murtagh, 2010b) is underpinned by the object set 
being embedded 
in  an ultrametric topology.  

We find that, in this framework, the information 
content is defined from the size of the object set, and not from any given 
object.
To that extent, therefore, the computational (or generative) potential
of unconscious thinking is far more powerful that that of conscious thinking.

\section{Text Analysis as a Proxy for Both Facets of Bi-Logic}  
\label{secttext}

Both conscious or asymmetric reason, and unconscious or symmetric 
reason, are facets of bi-logic according to Matte Blanco.  What
he means is that both play a role at different times, that these
roles are often complementary, and that the interplay of the two 
separate domains can be very revealing and instructive.

In this section we address the plausibility of 
appreciable
analysis of content of thought processes
based on interrelationships that in turn are 
frequencies of co-occurrence data.  Text will be used as
a proxy of underlying thinking, reasoning, conscious phenomena and 
also, every bit as much, representative of 
the underlying emotional, dreaming, or other 
unconscious mental processes.  
What we are seeking is an approach that is
 deployable and hence usable in practice.

Words are a means or a medium for getting at the substance and
energy of a story, notes McKee (1999, p.\ 179).
Ultimately sets of phrases express such underlying issues
(the ``subtext'', as expressed by McKee) as
conflict or emotional connotation.  
Change and evolution are inherent to a
plot.  Human emotion is based on particular transitions in thought.
So this establishes well the possibility that words and phrases
are not only taken literally
but can appropriately capture and represent such transition.
Text, says McKee, is the ``sensory surface'' of a work of art
(counterposing it to the subtext, or underlying emotion or
perception).

Simple words can express complex underlying reality.
Aristotle, for example, used words in common usage to express
technically loaded  concepts (Murtagh, 2005a, p.\ 169), and Freud did
also.

Rayner (1995) notes the following: ``The unconscious largely deals
not with particular logically asymmetrically locatable subjects and 
objects, but with abstract attributes, qualities or notions.  Put in 
another way, these propositional functions are adjectival and 
adverbial; they lie behind verbal nouns: lovingness, frighteningness
and so on.''  Such words, he notes, are ``abstract class attributes,
notions or conceptions'' and ``are the equivalent of the propositional
functions of the class''.   

This has an immediate bearing on the words used in  
unconscious processes.   Rayner (1995) notes the ``propositional
functions or abstract attributes'' or ``predicate thinking'', that
underly the unconscious as fundamental constituents.  He also 
briefly exemplifies this through  clinical work in schizophrenia and
child abuse by adults.  

One could of course deal with other units of thinking, or reasoning,
or unconscious processes, other than through words.  Chafe (1975), 
in relating and establishing mappings between memory and story,
or narrative, considered the following units.

\begin{enumerate}
\item {\em Memory} expressed by a {\em story} (memory takes the form of an
``island''; it is  ``highly selective''; it is a ``disjointed chunk''; but
it is not a book, nor a chapter, nor a continuous record, nor a stream).
\item {\em Episode}, expressed by a {\em paragraph}.
\item {\em Thought}, expressed by a {\em sentence}.
\item A {\em focus}, expressed by a {\em phrase} (often these
phrases are linguistic ``clauses'').
Foci are ``in a sense, the basic units of memory in that they represent 
the amount of information to which a person can devote his central 
attention at any one time''.
\end{enumerate}

The ``flow of thought and the flow of language'' are treated at once,
the latter proxying the former, and analyzed in their linear and
hierarchical structure as described in other essays in the same 
volume as Chafe (1979) and in Chafe (1994).


In the companion article to this article, Murtagh (2012), we address 
the following: Can we attempt to separate out good proxies for 
symmetrical logic and for asymmetrical logic?  To do this, we take
a great number of texts, relating to literature, technical writing,
and after-the-fact reporting on unconscious thought processes.

\section{Conclusion: Matte Blanco's Symmetric Logic as Thought 
Processes in an Ultrametric Information Space}
\label{sect8}

Ultrametricity, notes Lauro-Grotto (2007), can ``be used as a 
means to generate mental 
representations that hold in their inner structure all the 
contradictory aspects of experience and present a smooth surface 
allowing a kind of `easy handling' by mental processes.''  

Chapter 8 of Khrennikov (1997) deals with p-adic dynamical systems
in biology and social science.  Section 6 is entitled ``The human
subconscious as a p-adic dynamical system''.  The conscious is seen
as controlling the ``gigantic dynamical system'' that is the 
subconscious.  
A model of the unconscious is set up, based on ideas, that are
hierarchically related.  Dynamical systems on p-adic number encodings
of hierarchies are discussed, including how disruptive or ``manic 
ideas'' can be considered in such a context. 

This work on p-adic dynamical systems is taken further in the direction 
of application to cognitive processing in Khrennikov (2004).  (See 
also Khrennikov, 2007, and Khrennikov, 2010.) The 
author sets out to develop mathematical models for consciousness and 
allied or analogous mental processes, along the lines of what Newton, 
Descartes and others started in physics.  
Right from the start it is noted that ``human thinking (as well as many
other information processes) is fundamentally a hierarchical 
process''. 
An m-adic 
number encoding is used, expressing hierarchy and encompassing 
therefore an ultrametric topology.  Mental spaces, of ideas, are 
at issue.  A p-adic arithmetic, the author notes, is used as a ``mind
arithmetric''.  
A particular outcome of this work is the linking, by deployment of the
same mathematical approaches, of the mathematical 
models to quantum physical models.   

In chapter 7, dealing with ``abstract ultrametric information spaces'',
Khrennikov (2004) enunciates two conjectures that are compatible with the
exploration of the ultrametric topologies of ideas that are studied: 

\begin{enumerate}
\item ``Cognitive systems (at least some of them) are able to operate
simultaneously on all levels of the infinite cognitive hierarchy.''  
\item ``Consciousness is created by this infinite volume of information 
which is concentrated in a finite domain of physical space.'' 
\end{enumerate}

In a theorem (due to A. Lemin) on isometric embedding of an 
ultrametric space in a Euclidean space, Khrennikov notes that the 
theorem ``might be interpreted as the evidence of impossibility of
`spatial localization of mind' in brain.'' 


There is a clearly a great deal of compatibility between Matte Blanco's 
work and the other work that we discuss in this article.  

Finally we draw a link with symmetry and hierarchy, 
both understood in quite general terms.  
Herbert A.\ Simon, Nobel Laureate in Economics, 
believed in hierarchy at the basis 
of the human and social sciences, as the following quotation shows:
``... my central theme is that complexity frequently takes the form
of hierarchy and that hierarchic systems have some common properties
independent of their specific content.  Hierarchy, I shall argue, is
one of the central structural schemes that the architect of complexity
uses.'' (Simon, 1996, p.\ 184). 

\end{document}